# Learning a collaborative multiscale dictionary based on robust empirical mode decomposition

**Rui Chen,[a] Huizhu Jia,[a,*] Xiaodong Xie,[a] Wen Gao[a]**

[a]Peking University, National Engineering Laboratory for Video Technology, No. 5, Yiheyuan Road, Beijing, China, 100871

**Abstract**. Dictionary learning is a challenge topic in many image processing areas. The basic goal is to learn a sparse representation from an overcomplete basis set. Due to combining the advantages of generic multiscale representations with learning based adaptivity, multiscale dictionary representation approaches have the power in capturing structural characteristics of natural images. However, existing multiscale learning approaches still suffer from three main weaknesses: inadaptability to diverse scales of image data, sensitivity to noise and outliers, difficulty to determine optimal dictionary structure. In this paper, we present a novel multiscale dictionary learning paradigm for sparse image representations based on an improved empirical mode decomposition. This powerful data-driven analysis tool for multi-dimensional signal can fully adaptively decompose the image into multiscale oscillating components according to intrinsic modes of data self. This treatment can obtain a robust and effective sparse representation, and meanwhile generates a raw base dictionary at multiple geometric scales and spatial frequency bands. This dictionary is refined by selecting optimal oscillating atoms based on frequency clustering. In order to further enhance sparsity and generalization, a tolerance dictionary is learned using a coherence regularized model. A fast proximal scheme is developed to optimize this model. The multiscale dictionary is considered as the product of oscillating dictionary and tolerance dictionary. Experimental results demonstrate that the proposed learning approach has the superior performance in sparse image representations as compared with several competing methods. We also show the promising results in image denoising application.

*Corresponding author: Huizhu Jia, E-mail: hzjia@pku.edu.cn.

## 1. Introduction

Sparse representations over redundant dictionaries have received a lot of attention in recent years. It is reasonable due to the fact that an important variety of natural images can be well approximated by a sparse linear combination of basis elements called atoms [1]. The atoms are usually chosen from some redundant basis, usually called a dictionary. Particularly, this overcomplete dictionary consists of the atoms whose number greatly exceeds the dimension of the image space. Compared with other combinations of atoms, images using sparse representation enable more flexibility to adapt the representation to the data. Thus, it can provide high performance and successful lead to state-of-the-art results, especially for a wide spectrum of image processing and computer vision applications such as image denoising, image compression, image restoration as well as image classification [2-5].

Sparse modeling aims to design or learn an appropriate dictionary to represent a given set of training signals as sparsely as possible. Several algorithms have been developed for this task, e.g. the K-SVD [6] and the method of optimal direction (MOD) [7]. Most approaches to dictionary learning can be divided into two categories: the analytic approach and learning based approach. In the analytic approaches, a mathematical model of the data is formulated as an implicit dictionary where atoms are generated by various transforms such as discrete cosine transform (DCT) and wavelets [8-10]. The dictionaries generated by these approaches are highly structured and have fast implementation. In contrast, the learning-based approaches infer nonparametric dictionaries from a set of training data. However, the performance of such approaches hinges on an ideal dictionary for signal representation [11]. In general, the learned dictionaries have the potential to fit the given signals better than the analytic dictionaries [12].

Because natural images have essentially sparse distributions, most of important information in an image can be well preserved in multi-components or transform coefficients by employing elaborated signal analysis tools [13]. The effectiveness of multiscale representations for the specific images is highly dependent on how efficiently the constructed transforms can sparsify images of interest. Many types of multiscale decomposition approaches including the framelet, ridgelet, contourlet, curvelet and other wavelet frames [4,13,14] have been proposed to apply in various image processing tasks. Nevertheless, these image analysis tools fail to optimally self-adjust according to the characteristics of signal itself. To adaptively handle non-stationary and non-linear signals, Huang et al. [15] introduced the empirical mode decomposition (EMD) technique which decomposes the signal into a collection of intrinsic mode functions (IMFs) without any pre-determined basis functions. It consists of a local and fully data-driven separation of a signal in fast and slow oscillations plus a final monotonic or constant trend. Essentially, each IMF can be seen as an amplitude modulated-frequency modulated (AM-FM) oscillatory basis function in sense of compressive sensing theory [16]. Unlike the conventional scale-estimation methods, EMD is now considered as a powerful tool for fully adaptive multiscale image analysis and for quantifying monotonic components at the intrinsic scale level [17]. By advantage of good properties of EMD framework, this scheme can be extended for multiscale image representations.

*1.1. Related work*

There has been a great deal of efforts to merge the ideas of multiscale image representations with dictionary learning and sparse coding [18-26]. One line of work takes advantage of wavelet coefficients or pyramid decomposition to design the dictionaries. Ophir et al. [18] proposed to learn sub-dictionaries at different wavelet bands, which have left the atoms with the spatial correlation between wavelet coefficients, and the overall multiscale dictionary is able to produce a much efficient and effective representation than wavelet and K-SVD. Yan et al. [19] employed the multiresolution structure and sparsity of wavelets to learn a nonlocal dictionary in each decomposition level. Hughes et al. [20] decomposed an input image into a pyramid of distinct frequency bands, and then dictionary learning is performed on individual levels of the resulting pyramid where a set of atoms are learned across all scales. Especially, some methods combine the multiscale traits of both transform operator and dictionary itself. Mairal et al. [21] used a multiscale quadtree model to decompose big image patch along the tree to patches of small scales, and then a dictionary is learned at each scale. By advantage of joint sub-dictionaries for all sub-bands of stationary wavelet, Yin [22] presented to learn a multiscale dictionary for capturing the intrinsic characteristics of images. Chen et al. [23] developed two dictionary learning methods including cross-scale cooperative learning and cross-scale atom clustering to characterize the pattern similarity and uniqueness of designed dictionary in different scales, and then a cross-scale matching pursuit algorithm is employed for multiscale sparse coding. Another line of work focuses on the synthesis of analytic sparsifying dictionary and adaptive learning dictionary. Rubinstein et al. [24] presented a flexible sparsity model of the dictionary atoms over a fixed base dictionary to enhance the adaptability and compact representation. Ravishankar et al. [25] introduced the idea of learning doubly sparse transforms for analyzing the signals or image patches, where the square transform is taken as a product of a fixed analytic transform and an adaptive matrix constrained to be sparse. Hawe et al. [26] proposed a separable structure on the learned dictionary which permits large patch size and can be applied efficiently in reconstruction tasks. Moreover, during the training multiscale dictionary stage, some algorithms attempt to explicitly enforce the basic dictionary properties such as local correlation and mutual coherence. For instance, the work reported in [27] proposed to deal with learning dictionary by adding the decorrelation step in order to reach a target mutual coherence level.

Although it has proved to be reliable and suitable for image structural analysis, the standard EMD technique still has some drawbacks such as the sensitivity to noised data, intermittences and mode mixing [16]. Numerous researches on the EMD theory have been carried out essentially for the performance improvement. So taking the developments in joint signal scale-space representations, many modified EMD approaches have been proposed to alleviate the limitations [28-34]. Moreover, it could be extended to analyze image data with a superposition of a reasonably small number of approximately harmonic components. In order to remove the noisy components while retaining signal components of interest, the ensemble EMD (EEMD) [29] is developed to perform the EMD over an ensemble of the signal plus Gaussian white noise by using the statistical characteristics of white noise, and then the final EEMD is the average of each EMD. In addition, several works have extended the original EMD to 2D formulation through improving extrema detection, criteria for stopping the iterations of each IMF and interpolation techniques [30,31]. Bi-dimensional EMD (BEMD) [29] has been proposed to analyze image data. In this method, the rows and columns are independently decomposed with the standard EMD, and then the results are combined based on minimal scale strategy [32]. The recent optimization approaches for EMD have shown better performance in decomposing images into AM-FM components such as the low temporal complexity and orientation independence. Huang et al. [33] constructed a sparse data-adapted basis for input image to extract important features, where a smoothness functional is optimized by imposing the desired constraints of the resulting EMD involving the extrema and envelopes. Oberlin et al. [34] optimized a mean function to search for a local mean and reduce the mode-mixing phenomenon, in which the amplitudes of the modes of image data are added some symmetry constraints and the explicit interpolations through all the extrema are replaced. Colominas et al. [35] introduced an unconstrained optimization approach to find the AM-FM decomposition of images while making no explicit use of local smooth and extreme properties.

Compared to the existing multiscale learning methods, our approach is able to more effectively learn a multiscale compact dictionary. Concretely, our learned dictionary has strong representative capabilities to overcome the known weaknesses. Own to excluding the disturbing factors at image decomposition and dictionary learning stages, our scheme is robust to the noise and outliers. The improved EMD can produce fully adaptive and stable scales according to image content instead of pre-defined basis. Furthermore, through analyzing the clustered IMF basis of the given samples, a set of basis at different spatial scales can be selected to ensure the constructive stability for large-scale samples. By taking full advantage of adaptivity inherited in the EMD, the dictionary size could be automatically determined. Moreover, other important structure parameters involving the amplitudes and frequencies are also reliably known.

*1.2. Our work and contributions*

In this work, we propose to construct a novel multiscale dictionary learning framework based on a robust EMD. The main idea is to look for the sparsest multiscale representations of images within the largest possible dictionary consisting of IMFs. Our multiscale image decomposition method is derived from EMD based on structural characteristics of image contents. Moreover, it can be implemented very efficiently. In our EMD algorithm, the envelopes are estimated by optimizing a smoothness functional, and the mode mixing in sifting process is resolved with adaptive masking. Due to the fact that this improved EMD is robust to noisy and intermitted signal, natural images are reliably represented. At decomposition stage, the generated AM-FM atoms are used to construct an IMF dictionary. By characterizing this set of atoms, we can effectively exploit the local dependency and coherence among image structures. To promote the compact coding, the atoms are further selected via frequency clustering. In addition, a tolerance dictionary is learned to tune the sparsifying level and extend the generalization by solving the coherence regularized functional. A fast proximal scheme introduced for learning such dictionaries involves the closed-form solutions. A theoretical analysis provides the convergence guarantee for this scheme. We have analyzed and compared sparsely representative performance of the proposed learning algorithm. The experimental results have shown that our method produced state-of-the-art denoising results at different noise levels.

To be more specific, the main contributions of our work are given below:

1) We propose an improved EMD for dictionary learning. Unlike other EMD methods, our approach is robust to the noise and outliers because the bilateral filter ensures to reduce the noise and the optimization scheme can enhance the estimate accuracy of oscillating components. This robust EMD is well suited for multiscale image decomposition.

2) We use all the IMFs to construct a raw multiscale dictionary based on the robust EMD. In order to obtain compact representation, the typical atoms are further selected via frequency clustering. This refined IMF dictionary can adaptively reflect intrinsic structures of image itself with optimal sizes. Moreover, a tolerance dictionary is learned to enhance the sparsity level and generalization capability. The product of an IMF dictionary and a tolerance dictionary is acted as the final multiscale dictionary.

3) We present a coherence regularized dictionary learning model to jointly minimize the data approximation error and the coherence of atmos. A fast proximal scheme is developed to derive the optimization solutions for the proposed model. The resulting dictionary, equipped with the multiscale semi-structured atoms, enables us to obtain more effective and efficient sparse representations whereas requiring a lower computational time.

The rest of paper is organized as follows. In Section 2, the robust EMD is introduced and the implementation is given in detail. The proposed multiscale dictionary learning framework is presented in Section 3. The corresponding numerical optimization scheme is given in Section 4. The experimental results are illustrated in Section 5, and finally we conclude and discuss this work in Section 6.

## 2. Sparse representation using a robust EMD

Inspired by the advancements in analytic performance, we specifically design robust EMD method to sparsely represent image components when both the noise and intermitted points are present. The basic idea behind this approach is not only the accurate removal of local median tendency from a signal but also the reduction of very similar oscillations in different modes by means of a sifting processing. Specially, one tries to decompose a signal as the sum of a local median and an IMF sequentially. This sifting process continues until we obtain a satisfactory

IMF. The generated IMFs with a finite number are nearly orthogonal to each other. This method is very stable to noise perturbation and outliers. The details are introduced in this section.

*2.1. Mean envelope construction using bilateral filter*

In iterative decomposition procedure, the mean envelope is defined as the average of the upper and lower envelopes, which are respectively generated by interpolating the extracted local maxima and minima points in the image [15,29]. Because main oscillatory patterns of an image are confined between two envelopes, computing a good mean envelope is crucial to the success of the EMD type algorithm [32]. Moreover, the accuracy of the envelopes in terms of shape and smoothness is very important, which calls for the need to identify an appropriate 2D scattered data interpolation technique. In most EMD-based methods, all the extrema are identified by comparing all noisy data in segment lines and interpolated with cubic spline [31]. The BEMD algorithm is to perform the extrema extraction using the morphological filter in sliding windows [33]. Due to the instabilities to the noise and outliers, these methods are difficult to obtain the mean envelope accurately. This result will cause the deviation of tendency estimation for the AM-FM image structures and the wrong separations of its components [34].

The key stage in our EMD procedure is first to smooth fine-scale oscillating components while retaining the IMF information in the image $\mathbf{I}$. The bilateral filter is a well-known technique to remove the noise and smooth an image while preserving fine features [30]. It is a 2D weighted averaging filter with the closeness smoothing function in terms of spatial distance and the similarity weight function in terms of intensity difference. We adapt this filter to process image data while well balancing between the smoothing mechanism and the preserving capability. Each point in expected envelope $\mathbf{E}$ is smoothed by an implicit filtering process. The estimated output value $\hat{\mathbf{E}}[k,j]$ at every position is evaluated by

$$\hat{\mathbf{E}}[k,j] = \sum_{m=-H}^{H} \sum_{n=-H}^{H} w_{\{m,n\}}[k,j]\mathbf{E}[k-m,j-n] \qquad (1)$$

where the index $H$ dictates the neighboring range between the center pixel $(k, j)$ and its neighbors $\{(k-m, j-n); m, n \neq 0\}$. The weights are computed by

$$w_{\{m,n\}}[k, j] \propto \exp\left(-\frac{\Delta I[m,n;k,j]^2}{2\sigma_r^2} - \frac{m^2+n^2}{2\sigma_s^2}\right) \quad (2)$$

The function $\Delta I[m,n;k,j]$ denotes the difference between the intensities of pixel pairs. Considering for the smoothing effect, $H$ is set to be small value. In addition, the important parameters of this filter are adaptively tuned rather not pre-determined. The parameters, $\sigma_r$ and $\sigma_s$, are locally adaptive to the noise and image content. In detail, $\sigma_r$ is set to the same value of size of image patch, and $\sigma_s$ is set as the value close to the standard deviation $\sigma$ of noise present in the image.

The local extrema are used to detect the oscillations at different scales. By interpolating a set of extremal points $\mathcal{E}$, the maxima and minima envelopes can be constructed. And then the mean envelope is also obtained by averaging them. By imposing local smoothness and correlation constraints, we seek the upper and lower envelopes by optimizing the following functional

$$\arg\min_{\mathbf{E}} \left\{\|\mathbf{P}_e(\mathbf{E}-\mathbf{I})\|_2^2 + \lambda\|\mathbf{E}-\hat{\mathbf{E}}\|_2^2\right\} \quad (3)$$

where the matrix $\mathbf{P}_e$ depends on the locations of local extrema and it only imposes on the mode $\mathbf{E}[k_e, j_e] = \mathbf{I}[k_e, j_e]$ at each extrema $(k_e, j_e) \in \mathcal{E}$. So the positions of unit entities in the matrix $\mathbf{P}_e$ can be determined. The regularization parameter $\lambda > 0$ is used to control the trade-off between the smoothness and IMF-like mode. Based on the constructed relationship in (1), the quadratic functional (3) can be reduced to solving a sparse linear system. This convex optimization problem has a unique solution:

$$\mathbf{E}^* = [\mathbf{P}_e^T\mathbf{P}_e + \lambda(\mathbf{D}-\mathbf{W})^T(\mathbf{D}-\mathbf{W})]^{-1}\mathbf{P}_e^T\mathbf{P}_e\mathbf{I} \quad (4)$$

Here, $\mathbf{D}$ is a diagonal matrix and its diagonal elements are set as 1. The entries of the weighting matrix $\mathbf{W}$ are computed according to the Eq. (2). When the entries are out of the border of the matrix $\mathbf{W}$, we simply set their values to zero. The mean envelope is obtained by averaging the different matrices $\mathbf{E}^*$.

## 2.2. Sifting process based on adaptive masks

The sifting process aims at decomposing the image into superpositions of a small number of intrinsic modes that contain the components symmetrically oscillating around zero and the final monotonic trend [9]. The IMF is iteratively identified from the fine to coarse scales. One then decides whether or not to accept the obtained IMF that must satisfy two conditions: 1) In the whole data set, the number of extreme points (local maxima and minima) and the number of zero-crossing points must be equal to each other or differ by one at most; 2) At any point of the image, the mean value of the envelope defined by the local maxima and the envelope defined by the local minima must be equal to zero. If an image contains the intermittencies, the generated $i$-th IMF $h_i$ might introduce the mode mixing problem, for which the aliasing effect would exist and the components have at least two different frequencies [33]. Based on the sampling theory, we identify the mode mixing in $h_i$ by using the following criteria:

$$0.5\ell_{min} < \bar{\ell} < 2\ell_{max} \tag{5}$$

where the parameters $\ell_{min}$ and $\ell_{max}$ denote the distances with minima and maxima values in the set $\mathcal{L}$, respectively. Its element is computed by the distance between any adjacent extrema. In this set, $\bar{\ell}$ is the element with the median value. If this condition is satisfied, it manifests that the mode mixing exists in the candidate IMF $h_i^c$.

The masking signal technique has been adopted to solve the mode mixing problem of an IMF [28]. Based on this technique, a cosine signal is constructed as the mask $\mathbf{M}_{cos}$ and then added to input image $\mathbf{I}$. This adaptive masking technique can reduce the lower frequency components included in the candidate IMF. We select the median frequency determined by $\bar{\ell}$ as the masking frequency. Thus, the components with modulated frequencies are effectively separated. This adaptive mask is formulated as

$$\mathbf{M}_{cos}[k, j] = \bar{A}_M \cos \frac{2\pi}{\bar{\ell}} k \cos \frac{2\pi}{\bar{\ell}} j \tag{6}$$

where the amplitude $\bar{A}_M$ is computed as the arithmetic average of the maximum and minimum amplitudes of $h_i^c$. The auxiliary IMFs $h_i^+$ and $h_i^-$ can be obtained by sifting operation on $\mathbf{I}+\mathbf{M}_{cos}$ and $\mathbf{I}-\mathbf{M}_{cos}$. The IMF $h_k$ is generated by averaging $h_i^+$ and $h_i^-$.

*2.3. Raw sparse representation*

Based on the above improvements for original EMD technique, the whole implementation steps of the proposed EMD scheme are described in details as follows:

---
**Algorithm 1:** Robust EMD for Image Representation
---
**1: Input** the image $\mathbf{I}_{(i)}[j,k]$ with size of $M \times N$, and set $\sigma_r = 11$, $\sigma_s = \sigma$, $H = 5$, $\lambda = 0.8$, $i = 1$.
**2: Identify** all points of 2D local maxima and all points of 2D local minima of $11 \times 11$ image patches.
**3: Smooth** the estimated envelope by using the Eq. (2) to compute the weights at every point.
**4: Create** the maxima envelope $E_{max}$ and the minima envelope $E_{min}$ by solving the Eq. (4).
**5: Compute** the mean envelope $\bar{E}_{(i)} = (E_{max} + E_{min})/2$.
**6: Extract** the candidate IMF function $h_i^c = I_{(i)} - \bar{E}_{(i)}$.
**7: Repeat** the steps 3 to 6 until it can satisfy two IMF conditions, and then set IMF $h_i = h_i^c$.
**8: Detect** mode-mixing problem for $h_i$ using the Eq. (5). If exists, the masking signal $\mathbf{M}_{cos}$ is constructed using the Eq. (6). Then compute $h_i^+$ and $h_i^-$ to update the obtained IMF as $h_i = (h_i^+ + h_i^-)/2$.
**9: Estimate** the residual component $\mathbf{R}_{es} = \mathbf{I}_{(i)} - h_i$; If it does not contain any more points, the EMD process is stopped; Otherwise, update $i = i+1$ and $\mathbf{I}_{(i)} = \mathbf{R}_{es}$, and then go to step 2.
**10: Output:** the final IMFs $\{h_i\}$ and the residue $\mathbf{R}_{es}$.
---

By means of our robust EMD method, the generated IMFs can adaptively form a complete and nearly orthogonal basis for representing 2D signal. Then the given image $\mathbf{I}$ is decomposed with the following formulation:

$$\mathbf{I}[j,k] = \sum_{i=1}^{Q} h_i[j,k] + \mathbf{R}_{es} \qquad (7)$$

where the superscript $Q$ is the total number of 2D IMFs and it represents the scale level for image decomposition. $\mathbf{R}_{es}$ denotes the residue of the image decomposition and is generally a small monotonic or constant component. The IMFs with lower indices contain the faster oscillations present in the image whereas the IMFs with higher indices contain the slower oscillations. We adopt these IMFs as the atoms to construct an overcomplete base dictionary. In practice, accounting for the repetition patterns of image patches, each IMF $h_i$ is further partitioned into, arbitrary number $r$, IMF-type atoms $\{\mathbf{d}_1^i,\cdots,\mathbf{d}_r^i\}$, which can be typically approximated as the form $A(j,k)\cos\varphi(j,k)$. As a result, the reconstructed image $\hat{\mathbf{I}}$ is sparsely represented using a raw dictionary $\mathbf{D}_0$:

$$\begin{cases} \hat{\mathbf{I}} = \mathbf{D}_0 \alpha, \ \mathbf{D}_0 = [\mathbf{d}_1^1,...,\mathbf{d}_r^1\,|\,...\,|\,\mathbf{d}_1^Q,...,\mathbf{d}_r^Q] \\ \mathbf{d}_i \propto A_i(j,k)\cos\varphi_i(j,k) \quad \forall i \end{cases} \quad (8)$$

where the coefficient vector $\alpha$ essentially affects the reconstruction quality of image structures. The IMF dictionary $\mathbf{D}_0$ is constituted by atoms $\{\mathbf{d}_i\}_{i=1}^{Qr}$. In this coding paradigm, raw spare approximation is adaptively found by dividing each 2D IMF into small atoms while preserving multiscale image features.

## 3. Multiscale learning algorithm via cross-scale collaborative dictionaries

An efficient dictionary is able to maximize the sparsity level and minimize the reconstruction error of image representation. In an effort to further promote sparsity of a multiscale setup, we propose a new dictionary learning scheme via the collaboration between an IMF dictionary and a tolerance dictionary. The mutual coherence of across-scale atoms are taken as prior information to reduce the scale mixing. The details are explained in this section.

*3.1. Atom selection via IMF clustering*

Each atom $\mathbf{d}_i$ in dictionary $\mathbf{D}_0$ conforms to the IMF condition, which can measure multiscale structures of images in terms of the amplitudes, frequencies and phases. Essentially, AM-FM model provides a perceptually motivated image representation in which nonstationary amplitude and frequency information might be physically interpreted. The AM function is a local characterization of local contrast in the image. The FM function reflects a dense characterization of local texture orientations and pattern locations. Based on these important attributes, the raw dictionary $\mathbf{D}_0$ is possible to contain many atoms with high coherence. Therefore, this dictionary need be further sparsified at different scales. To obtain more compact dictionary, we select a set of principal atoms from $\mathbf{D}_0$. In our scheme, the significant frequencies and amplitudes of atoms are employed as the important criteria of this selection process. Compared to predefined atoms generated by DCT, the IMF atoms more accurately represent a single image or a group of images.

Based on the results of robust EMD, the amplitudes and frequencies of atoms can be obtained from the IMFs. Let $c_k^{d_i}$ denote the absolute difference between any two neighboring zero points of each IMF atom $\mathbf{d}_i$. We select the median value $c_M^{d_i}$ from $\{c_k^{d_i}\}_{k=1}^r$ to characterize the frequency. $c_{min}$ and $c_{max}$ are the minimal and maximal elements in the set $C=\{c_k^{d_i}\}_{i,k}$, respectively. For the atom $\mathbf{d}_i$, the amplitude $\bar{A}_i$ is computed by averaging all the absolute values of the distances from the extrema to zero axis. The atoms are first clustered according the similar frequencies. Given all representative atoms $\{d_i\}_{i=1}^{Qr}$, the degree of membership to each cluster is computed by

$$kc_{min} \leq c_M^{d_i} \leq (k+1)c_{min} \leq c_{max}, k \in \mathbf{Z}^+ \qquad (9)$$

According to the Eq. (9), all the atoms in $\mathbf{D}_0$ are divided into the different groups $\{G_j\}_{j=1}^N$. Let $\bar{A}_{max}^{G_j}$ be the maximum in the set $\{\bar{A}_i \in G_j, j < k+1\}$. Based on the fact the magnitudes of each component are tightly related with the Fourier spectra of natural images [36], we select the atoms with the maximal amplitudes $\{\bar{A}_{max}^{G_j}\}_{j=1}^N$ as each cluster centroid, where AM-FM components with typical frequency power represent principal structures in the image. Then we would employ these refined atoms to construct a new IMF dictionary $\tilde{\mathbf{D}}_0$. It can be concluded that this refined dictionary has high compactness which makes the learned atoms discriminative [37]. In fact, the dictionary size is automatically estimated and adaptive to the input data to describe the compactness of output atoms. By this means, all the best multiscale atoms consist in the dictionary $\tilde{\mathbf{D}}_0$ with a small reconstruction error.

### 3.2. Problem formulation

Given a matrix $\mathbf{Y} = [\mathbf{y}_1, \mathbf{y}_2, ..., \mathbf{y}_N] \in \mathbf{R}^{n \times N}$ whose columns represent training samples, learning an adaptively collaborative dictionary $\mathbf{D} \in \mathbf{R}^{n \times K} (K > n)$ with $K$ atoms for sparse representation of $\mathbf{Y}$ is typically considered as a joint optimization problem. With respect to the corresponding sparse coefficients $\mathbf{X} = [\mathbf{x}_1, \mathbf{x}_2, ..., \mathbf{x}_N] \in \mathbf{R}^{K \times N}$, this sparse coding problem for $\mathbf{Y}$ can be formulated as follows

$$\arg\min_{\mathbf{D},\mathbf{X}} \frac{1}{2} \|\mathbf{Y} - \mathbf{D}\mathbf{X}\|_F^2 \quad s.t. \quad \|\mathbf{x}_i\|_0 \leq p \ \forall i \tag{10}$$

Here, the notation $\mathbf{x}_i$ denotes $i$-th column of the matrix $\mathbf{X}$. The parameter $p$ represents the desired sparsity level for each training image. Although solving this non-convex problem is in general NP-hard, it can be solved exactly under certain conditions [38]. To scale down the feasible domain and find the optimal solutions, we propose to model the sparse structure of $\mathbf{D}$ as the formulation $\mathbf{D} = \mathbf{B}\tilde{\mathbf{D}}_0$, which admits the overall advantage of learning-based adaptivity and the efficiency of robust EMD. The square matrix $\mathbf{B} \in \mathbf{R}^{n \times n}$ is termed as a tolerance dictionary and its rows are rearranged as the atoms $\mathbf{B} = [\mathbf{b}_1, ..., \mathbf{b}_n]^T$. The effect of any atom $\mathbf{b}_i$ is to adjust local shape of the IMFs in the dictionary $\tilde{\mathbf{D}}_0$ aiming to strengthen the generalization and sparsification accuracy. Moreover, to penalize the large frequency and phase shifting of each IMF atom, the $l_0$ quasi norm constraint on matrix $\mathbf{B}$ enforces sparsity of the entries of the matrix $\mathbf{B}$. We assume $q$ non-zeroes in $\mathbf{B}$. Based on these analysis, we propose the following problem formulation for learning a tolerant sparse representation

$$\arg\min_{\mathbf{B},\mathbf{X}} \frac{1}{2} \|\mathbf{Y} - \mathbf{B}\tilde{\mathbf{D}}_0\mathbf{X}\|_F^2 \quad s.t. \quad \|\mathbf{B}\|_0 \leq q, \ \|\mathbf{x}_i\|_0 \leq p \ \forall i \tag{11}$$

When performing the sparse coding for Eq. (11), the learned dictionary $\mathbf{B}$ may contain similar atoms, thus leading to a degenerate dictionary. Mutual coherence of a dictionary measures the maximal correlation between any distinct atoms [3]. To avoid similar atoms appearing in the dictionary $\mathbf{B}$, we consider restricting its coherence $\mu(\mathbf{B})$. We define the coherence in a row-wise way as

$$\mu(\mathbf{B}) = \max_{\forall i,j, i \neq j} \left| \left\langle \frac{\mathbf{b}_i}{\|\mathbf{b}_i\|_2}, \frac{\mathbf{b}_j}{\|\mathbf{b}_j\|_2} \right\rangle \right| \qquad (12)$$

From this definition, we have $0 \leq \mu(\mathbf{B}) \leq 1$. If the coherence is close to 1, it means that there are very similar rows, which is the case we attempt to avoid. Besides, the trivial solutions including zero rows can be excluded. According to theoretical results on sparse coding, mutual coherence of dictionary has direct impact on stability and performance of coding algorithm. A lower coherence promotes better sparse recovery. Furthermore, reducing the coherence of the dictionary can increase its generalization ability and sparse coding performance.

With better bounded conditions, the extra unit norm constraint forces on each row of matrix $\mathbf{B}$ to eliminate scaling ambiguity and preserve signal shape. Then, introduced as an alternative the representation for the coherence, the Gram matrix $\mathbf{B}^T\mathbf{B}$ has unit values on the main diagonal and the values of off-diagonal elements are the inner product of two atoms. Thus, as an important property of the dictionary, the incoherence constraint on $\mathbf{B}$ can be transformed into the term $\|\mathbf{B}^T\mathbf{B} - \Lambda\|_F^2$, which measures the Frobenius distance between the Gram matrix $\mathbf{B}^T\mathbf{B}$ and the identity matrix $\Lambda$. The Gram matrix of an orthonormal dictionary corresponds to the zero mutual coherence. As a convex relaxation of $l_0$ norm for sparsity-inducing regularizer, we replace $\|\mathbf{B}\|_0$ by $\|\mathbf{B}\|_1$ to provide an approximate optimal solution. This relaxation can improve the computational feasibility and efficiency of sparse coding without loss of the tolerant requirements for dictionary $\mathbf{B}$. The penalty terms for $\mathbf{B}$ are used in objective function to restrict the correlations between the atoms. This is done by reduction of over-fitting to the training data and avoiding atom degeneracy. Equipped with these constraints, we reformulate Eq. (11) as the following minimization problem:

$$\begin{cases} \arg\min_{\mathbf{B},\mathbf{X}} \dfrac{1}{2}\|\mathbf{Y} - \mathbf{B}\tilde{\mathbf{D}}_0\mathbf{X}\|_F^2 + \alpha\|\mathbf{X}\|_0 + \dfrac{\beta}{2}\|\mathbf{B}^T\mathbf{B} - \Lambda\|_F^2 + \gamma\|\mathbf{B}\|_1 \\ s.t. \ \|\mathbf{b}_i\|_2 = 1, \ 1 \leq i \leq n \end{cases} \qquad (13)$$

where $\alpha$, $\beta$ and $\gamma$ are the trade-off parameters that balance the data fitting term and three regularization terms. Two convex constraints on $\mathbf{B}$ help reduce both the sparsification error and marginal loss in the image representation. Note that enforcing more constraints on $\mathbf{B}$ are too restrictive and the desired solutions will provide almost no improvement to recover the correct tolerant dictionary. To prevent the over-penalization on large elements of a sparse vector caused by nonsmooth $l_1$ norm, the nonconvex $l_0$ quasi norm on $\mathbf{X}$ holds the original formulation in (13).

## 4. Optimization

By globally minimizing the objective function, the meaningful solutions to the unknown $\mathbf{B}$ and $\mathbf{X}$ can be reliably found. A fast proximal splitting minimization scheme is designed to deal with the problem in (13) by alternating between the sparse coding and dictionary update steps. The proximal alternating linearized minimization (PALM) method has been proposed for solving a class of nonconvex and nonsmooth problems with strong convergence [39]. Based on this general proximal optimization framework, we further develop a modified proximal algorithm to achieve optimal performance for the specific problem (13). In dictionary update stage, the block coordinate descent (BCD) technique [40] is introduced to accelerate the convergence rate. Moreover, the frequency tolerance conditions for each IMF atom are checked to prevent the large shape shifting and residual phase in each iteration step.

*4.1. Sparse approximation*

With respect to the coefficient matrix $\mathbf{X}$, the feasible set can be found by employing the linearized proximal operator to directly solve the nonconvex $l_0$ norm optimization problem. For the sum of a finite collections including the proper lower semi-continuous functions and smooth functions with Lipschitz gradient on any bounded set [41], a proximal regularization is coupled into the objective function and then this composite function is assumed to satisfy the so-called Kurdyka-Łojasiewicz (KL) property for completely removing the convexity setting [39]. Building on the PALM, the proximal operators are taken on the nonconvex and nonsmooth parts to obtain descent iteration steps. The initial solution $\mathbf{x}^0$ can be simply chosen as the unit matrix. While fixing the multi-block dictionary variables $\mathbf{B}^{k,i}$, the sparse code $\mathbf{x}^k$ is updated by solving the following subproblem:

$$\begin{cases} \mathbf{X}^k \in \arg\min \dfrac{\theta^k}{2\alpha}\left\|\mathbf{X}-\mathbf{U}^k/\theta^k\right\|_F^2 + \|\mathbf{X}\|_0 \\ \mathbf{U}^k = \mathbf{X}^{k-1} - \dfrac{1}{\theta^k}\nabla H_{\mathbf{X}}(\mathbf{X}^{k-1},\mathbf{B}^{k,i}) \\ H_{\mathbf{X}} = \dfrac{1}{2}\left\|\mathbf{Y}-\mathbf{B}\tilde{\mathbf{D}}_0\mathbf{X}\right\|_F^2 \end{cases} \tag{14}$$

where $H_{\mathbf{X}}$ is a continuous and differentiable function. $\nabla H_{\mathbf{X}}$ is the partial gradient related with all the variables in matrix $\mathbf{X}$. The block-variables $\mathbf{B}^{k,i}=[\mathbf{b}_1^{k-1},\cdots,\mathbf{b}_{i-1}^{k-1},\mathbf{b}_i^k,\mathbf{b}_{i+1}^k,\cdots,\mathbf{b}_n^k]$ are updated according to the current values. The stepsize $\theta^k$ can be chosen as

$$\theta^k = \max\left\{\rho\left\|(\mathbf{B}^{k,i})^T\mathbf{B}^{k,i}\right\|_F, a\right\} \tag{15}$$

Here, the parameter $\rho > 1$ can affect the convergence rate and its value is set according to the Lipschitz modulis of function $H_{\mathbf{X}}$. The constant $a > 0$ is determined with the requirement of global convergence of the sequence $\{\mathbf{x}^k\}$.

Each iteration above requires solving the optimization problem (14). It can be decomposed into the summation of independent convex problems. The stationary solution of $\mathbf{x}^k$ exists for the bounded $\mathbf{X}$ and $\mathbf{B}$. The thresholding operator $T$ is used to decrease the degraded solutions. The minimization problem in (14) has the closed form solution given by

$$\begin{cases} \mathbf{X}^k = T_{\sqrt{2\alpha/\theta^k}}(\mathbf{U}^k) \\ T_{\sqrt{2\alpha/\theta^k}} = \max\{|\mathbf{U}^k|,\sqrt{2\alpha/\theta^k}\} \end{cases} \tag{16}$$

### 4.2. Dictionary update

Due to the existence of both norm equality and mutual coherence constraints on atoms $\{\mathbf{b}_i\}_{i=1}^n$, the subproblem for solving $\mathbf{B}$ is also a nonconvex problem. The proximal operator related with $\mathbf{B}$ has descent gradient and promotes convex in each block of variables. The dictionary variables $\mathbf{B}_i^k$ are updated cyclically by minimizing the objective with respect to one block of variables at a time while the other is fixed at its most recent values. Based on BCD scheme, the proximal mapping for nonconvex function can project each block into the bounded sequence $\{\mathbf{B}^k\}$ during each iteration. Herein, the detailed numerical scheme for solver of $\mathbf{B}$ is also a multi-block alternating iteration process. Since each atom in dictionary $\mathbf{B}$ requires the suitable deviation with the corresponding base atom in dictionary $\tilde{\mathbf{D}}_0$, the initial dictionary $\mathbf{B}$ is chosen as the unit matrix for a correct start. While holding the sparse code $\mathbf{x}^k$ fixed, the estimate of dictionary variables $\mathbf{B}_i^k$ can be solved by updating the following subproblem:

$$\begin{cases} \mathbf{B}_i^k \in \arg\min \left\| \mathbf{b} - \mathbf{S}_i^k \right\|_F^2 \\ \mathbf{S}_i^k = \mathbf{B}_i^k - \frac{1}{\delta_i^k} \nabla \mathbf{H}_\mathbf{B}(\mathbf{X}^k, \tilde{\mathbf{B}}^{k,i}) \\ \mathbf{H}_\mathbf{B} = \frac{1}{2} \left\| \mathbf{Y} - \mathbf{B}\tilde{\mathbf{D}}_0 \mathbf{X} \right\|_F^2 + \beta \sum_{i \neq j} (\mathbf{B}_i^T \mathbf{B}_j)^2 + \gamma \left\| \mathbf{B} \right\|_1 \end{cases} \quad (17)$$

where $\mathrm{H}_\mathbf{B}$ is a continuous and differentiable function. $\nabla \mathrm{H}_\mathbf{B}$ is the partial gradient related with all the variables in matrix $\mathbf{B}$. Note that the term $\left\| \mathbf{B}^T \mathbf{B} - \Lambda \right\|_F^2$ in (13) is equal to $2\beta \sum (\mathbf{B}_i^T \mathbf{B}_j)^2$ since the condition $\left\| \mathbf{b}_i \right\|_2 = 1$ is added. The block-variables $\tilde{\mathbf{B}}^{k,i} = [\mathbf{b}_1^k, \cdots, \mathbf{b}_{i-1}^k, \mathbf{b}_i^{k-1}, \mathbf{b}_{i+1}^{k-1}, \cdots, \mathbf{b}_n^{k-1}]$ are updated according to the current values. Because the Lipschitz constant of the function $\mathrm{H}_\mathbf{B}$ guarantees the convergence of $\{\mathbf{B}^k\}$ generated by iterates (17), the constant $b > 0$ is used to limit the stepsize based on the stationary conditions. The stepsize $\delta_i^k$ can be chosen as

$$\delta_i^k = \max \left\{ \rho(\left\| \mathbf{X}_i^k \right\|_2^2 + 2\beta \left\| (\tilde{\mathbf{B}}^{k,i})^T \tilde{\mathbf{B}}^{k,i} \right\|_F^2), b \right\} \quad (18)$$

According to the block-coordinate decreasing property of the proximal operator in (17), the solution of nonconvex part on $\mathbf{B}$ converges toward the stationary point. In addition, the update of component $\mathbf{B}_i^k$ at each iteration is checked to satisfy the correlation condition $\Pi_\mathbf{B}^k$ in order to prevent large IMF deviations. The closed form solution of problem in (17) is given by

$$\begin{cases} \mathbf{B}_i^k = \mathbf{S}_i^k / \|\mathbf{S}_i^k\|_2 \\ \Pi_{\mathbf{B}}^k = \dfrac{\left\langle \mathbf{B}_i^k \tilde{\mathbf{D}}_0^{(i)}, \tilde{\mathbf{D}}_0^{(i)} \right\rangle}{\|\mathbf{B}_i^k \tilde{\mathbf{D}}_0^{(i)}\|_2 \|\tilde{\mathbf{D}}_0^{(i)}\|_2} \geq \sqrt{2\beta/\tau} \end{cases} \quad (19)$$

where $\tilde{\mathbf{D}}_0^{(i)}$ denotes the vector corresponding to $\mathbf{B}_i^k$ when performing the matrix multiplication. The larger value of $\Pi_{\mathbf{B}}^k$ represents higher similarity between $\tilde{\mathbf{D}}_0^{(i)}$ and $\mathbf{B}_i^k$. If the inequality in (19) is not satisfied, $\mathbf{B}_i^k$ will terminate the self-update. The parameter $\tau$ adjusts the tolerance errors.

*4.3. Convergence analysis*

The proposed numerical optimization algorithm alternates between the sparse coding stage and the dictionary update stage with a strong convergence guarantee. Each subproblem is a nonsmooth and nonconvex minimization of the composite function with the inner function being linearized at the current iteration and the stepsize being updated to maintain a descent property. For a fixed dictionary, the exact $\mathbf{X}$ obtained in the analysis coding is the optimal solution under the constraints. Thus, the respective cost function (16) can only decrease in this stage. In the dictionary update stage, the update of $\mathbf{B}$ is along the descent direction and stepsize is chosen to guarantee that the update will not increase the cost function (19). The nonsmooth and nonconvex parts in (13) are the proper lower semi-continuous functions and thus have KL property. The proximal-gradient step in nonconvex setting can warrant sufficient decrease of the objective function. The iterative sequence $\{(\mathbf{X}^{(k)}, \mathbf{B}^{(k)})\}$ generated by our proximal algorithm for solving the nonconvex composite optimization problem is a bounded Cauchy sequence. Note that $H_{\mathbf{X}}$ and $H_{\mathbf{B}}$ are continuously differentiable functions. By the Lipschitz continuity of $\nabla H_{\mathbf{X}}$ and $\nabla H_{\mathbf{B}}$, the bounded sequence provably converges to a critical point $(\overline{\mathbf{X}}, \overline{\mathbf{B}})$ [39] of (13) at least in the sublinear convergence rate under the weak minima and this regular conditions. The proposed proximal method with a global convergence property is computationally efficient at solving each subproblem.

## 5. Numerical Experiments

We evaluate the performance of the proposed multiscale dictionary learning framework for image representation and image denoising. We first illustrate the multiscale representation effect for natural images by using our robust EMD. Then the collaborative learning capability is thoroughly analyzed and shown to improve. Furthermore, we compare our approach with several competing methods in terms of their ability to provide low sparsification errors and good denoising on a variety of natural images.

*5.1. Setup and implementation details*

The implementation of our learning method is coded in Matlab v7.10 (R2010a) and the computations of all methods in experiments are executed on a PC workstation with an Intel Core i7 CPU at 2.4 GHz and 8 GB memory. The sparsity level of the structured or unstructured dictionary is measured by the percent ratio between number of non-zeros and total atoms in the dictionary. The quality of sparse coding is judged based on normalized reconstruction error (NRE), which is defined as $\|\mathbf{Y} - \mathbf{DX}\|_F^2 / \|\mathbf{Y}\|_F^2$. When evaluating our method, $\mathbf{D}$ is set as $\mathbf{D}_0$ for the robust EMD, $\tilde{\mathbf{D}}_0$ for the refined IMF dictionary, and $\mathbf{B}\tilde{\mathbf{D}}_0$ for the overall learning quality. We evaluate the visual quality of the reconstructed and denoised images via two metrics including peak signal-to-noise ratio (PSNR) and structural similarity (SSIM) [42].

In robust EMD procedure, the parameter $\sigma_r$ is set to 11, and $\sigma_s$ is set as the same value of the standard variance $\sigma$ of Gaussian noise. The range $H$ is fixed at 5. The suitable value of $\lambda$ is set to 0.8 to obtain optimal solutions. At dictionary learning stage, the regularization parameters can essentially affect the accuracy of numerical solutions, and are reliably set to $\alpha = 150$, $\beta = 0.04$, $\gamma = 0.01$ uniformly for all experiments. To obtain proper stepsizes and fast convergence rate, the involving parameters are set as $\rho = 1.2$, $\tau = 0.14$, $a = 0.3$, $b = 0.5$. The maximum iteration number is set to 50 as a stopping condition. We would like to emphasize that our recommended parameters empirically perform well even with the addition of noise, and hence in practice does not require manual tuning.

*5.2. Effect of the robust EMD*

The important steps for the robust EMD (REMD) aim to analyze the frequencies, amplitudes, phases and directions by separating the intrinsic mean tendency and IMF patterns confined by extrema points. The first experiment is to evaluate the effect of main process of REMD. The standard *Barbara* image is decomposed by using Algorithm 1. Performing Eq. (4) for envelope construction, the mean image is computed and shown in Fig. 1(b). The first fine-scale IMF image is recursively extracted and given in Fig. 1(c). As seen from the corresponding scanline plots in Fig. 1(d), we have accurately detected the envelope $M_1$ meanwhile preserving the edges in $I$.

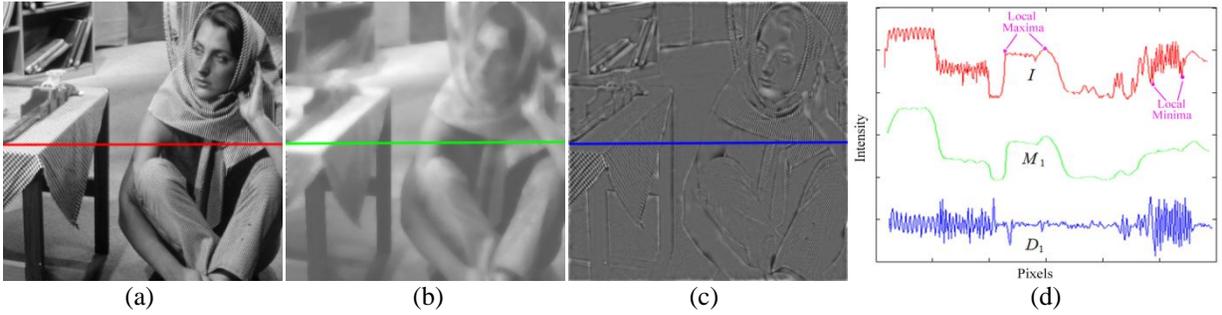

(a) (b) (c) (d)

**Fig. 1.** Main Decomposition process of REMD. (a) *Barbara* image. (b) Mean image. (c) IMF image. (d) Scanline plots of input intensities in main steps of REMD.

The REMD robustness for the noise and scale mixing is evaluated in the second experiment. Moreover, we compare our results with the state of the art method: UOA-EMD [35]. The input image in Fig. 2(a) is obtained from *Lena* image in Fig. 2(e) by adding the Gaussian noise with the standard deviation $\sigma = 20$. The decomposition results including two IMFs and the residues are shown in Fig. 2. We observe that the REMD works significantly better than the UOA-EMD and provides the best visual quality of decomposition results. Due to the effectiveness of our optimization model (4) incorporating the bilateral filter, the noise is removed in decomposition process and each IMF reveals the unpolluted textures in coherent scale. Moreover, the adaptive mask in Eq. (6) can decrease scale mixing and matching between the noise and image features.

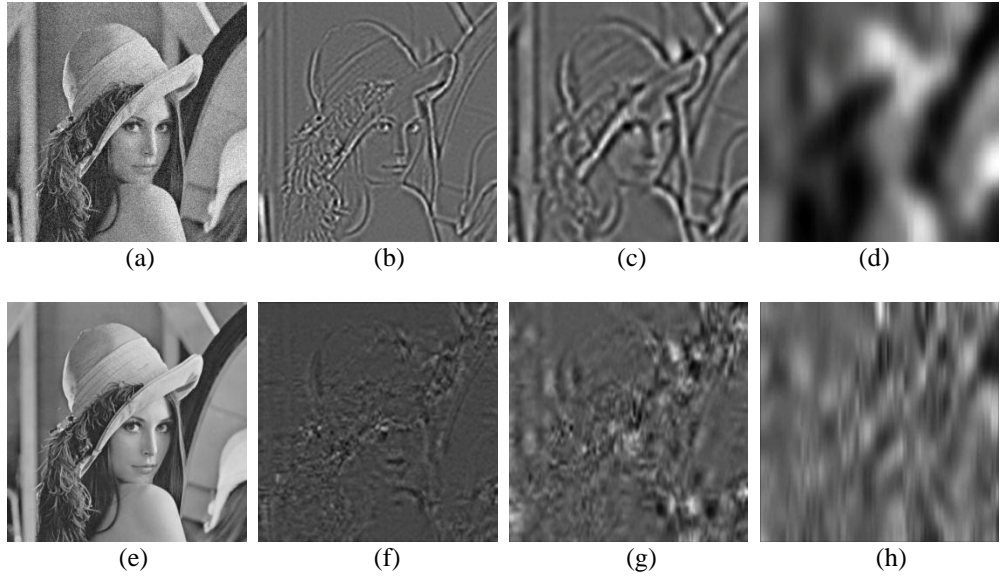

**Fig. 2.** Decomposition results for *Lena* image added the noise. (a) noisy image. REMD results: (b) 1st IMF, (c) 2nd IMF, (d) the residue. (e) original image. UOA-EMD results: (f) 1st IMF, (g) 2nd IMF, (h) the residue.

The overall REMD performance is quantitatively evaluated in terms of reconstruction errors and computational complexity. The $512 \times 512$ version of standard *Lena* image is used as the input. It can be seen from Fig. 3(a) that the REMD has smaller errors than the UOA-EMD when the noise added in the input is varied from low to high levels. Because the NRE values increase slowly and lie in the acceptable range, the REMD can reliably produce an IMF dictionary $\mathbf{D}_0$ from the noisy image. As shown in Fig. 3(b), all the multiscale IMFs for natural images are extracted usually in less than 30 iterations. Moreover, our REMD scheme has lower computation cost and the potential for the real-time applications.

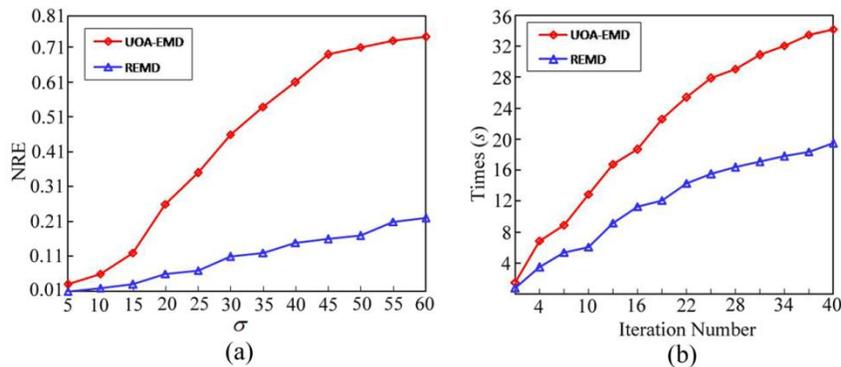

**Fig. 3.** Numerical analysis on the EMD methods. (a) NRE results. (b) computational times.

## 5.3. Evaluation on the multiscale dictionaries

The performance of two multiscale dictionaries $\tilde{\mathbf{D}}_0$ and $\mathbf{B}\tilde{\mathbf{D}}_0$ is extensively evaluated for important properties. In this experiments, all the raw IMF dictionaries $\{\mathbf{D}_0\}$ with respect to training images are refined based on the criteria of Eq. (9) and combined into one dictionary $\tilde{\mathbf{D}}_0$, in which the IMF atoms are sorted and ordered according to the amplitudes and frequencies. The quality of learned dictionary $\mathbf{B}$ in Eq. (13) is reflected via dictionary structure of $\mathbf{B}\tilde{\mathbf{D}}_0$.

The IMF atoms of size $32\times 32$ are shown in Fig. 4(a). Without the need of any initializations, these refined atoms are achieved by adaptation of the patches of *Barbara* image, and obviously exhibit regular geometric structures and texture-like features. Figs. 4(b) and (c) plot the average frequencies and amplitudes of IMF atoms whose distributions conform to the statistical results for natural images [36] and cover all important AM-FM components. Because the difference values of NRE in Fig. 4(d) increase slightly as a function of reduction number of clustered atoms, the refinement strategy in Eq. (9) is reasonable. Figs. 4(e) and (f) indicate that our scheme can provide better representations than the DCT at increasing patch size and sparsity level.

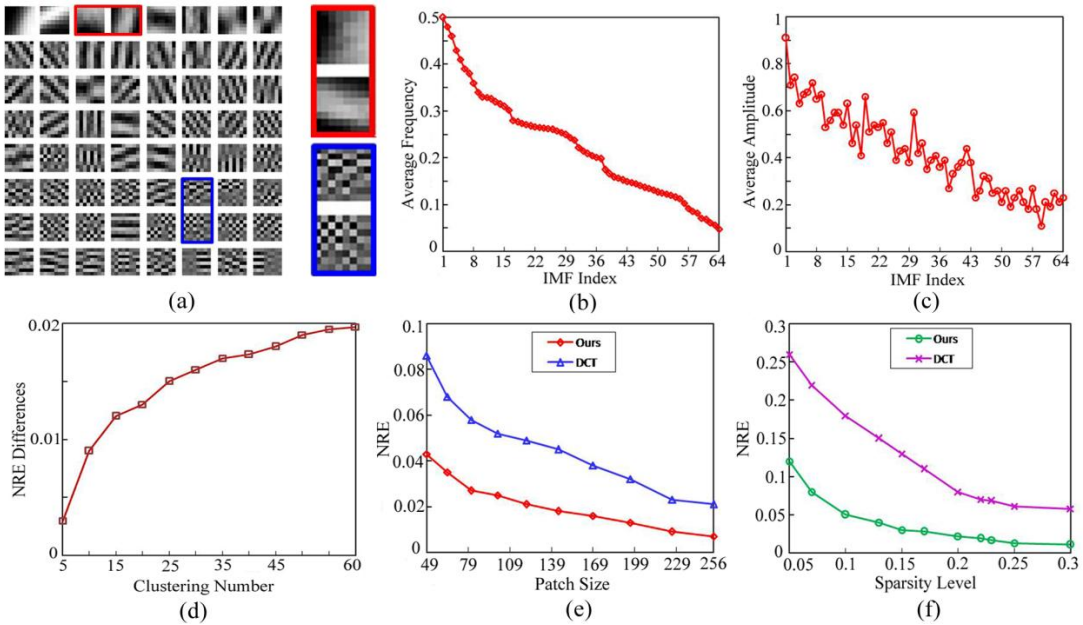

**Fig. 4.** Evaluation results on the refined dictionary $\tilde{\mathbf{D}}_0$.

By learning the tolerance dictionary from *Barbara* image, final IMF atoms in Fig. 5(a) are similar with the refined atoms, but the discrimination among pairwise atoms has improved much. In addition, important IMF parameters involving the frequencies and amplitudes are adjusted more reasonable according to the plots of Figs. 5(b) and (c). The explanations are that the global atoms improve the regularity for characterizing image features by finding optimal parameters. Observations from Fig. 5(d), most of atom pairs have small values of mutual coherence within the range [0.1, 0.3]. This result testified the effectiveness of decorrelation constraints in Eq. (13). When the number of training patches vastly increases, Figs. 5(e) and (f) show that global dictionary has compact structure and can obtain more accurate representation than the DCT.

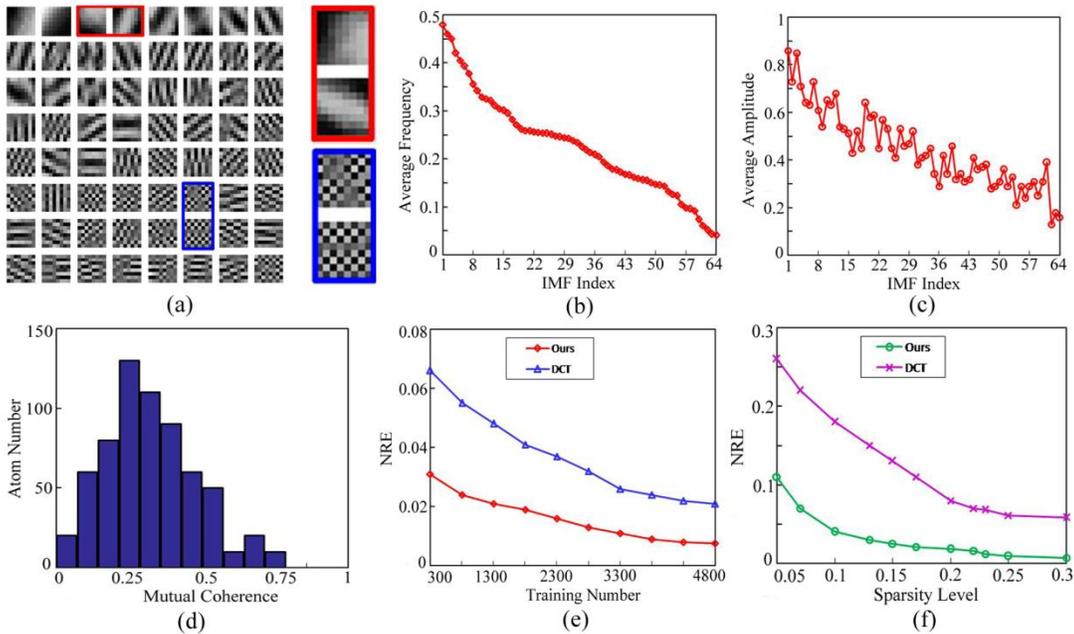

**Fig. 5.** Evaluation results on the global dictionary $B\tilde{D}_0$.

Although we considered the raw IMF dictionary $D_0$ with respect to a specific image, a refined version $\tilde{D}_0$ can be learnt over a class of images and tested on unrelated test images, and also provides better sparse representations than fixed transforms such as DCT. By means of the latent tolerance mechanics of sparse learning, such global dictionary $B\tilde{D}_0$ can further enhance the generalizability and meanwhile effectively inherits the optimal IMF dictionary structures.

*5.4. Application on image denoising*

We apply the proposed multiscale dictionary generation and learning framework to image denoising for demonstrating the performance of representing various images. Moreover, we compare the denoising results obtained by our algorithm with those obtained by two learning-based denoising methods including the K-SVD [2] and the OCTOBOS [43], and one state-of-the-art BM3D denoising method [42]. Note that the recommended optimal settings and implementation for three compared methods are adopted in our experiments. The goal of image denoising is to recover an estimated original image $\hat{\mathbf{Y}}$ from its measurement $\mathbf{Y}$ corrupted by the additive Gaussian noise with zero mean and variance $\sigma^2$. Similar to the typical dictionary-based denoising methods [2,43], we work with overlapping image patches. Based on the extracted IMF patches using the REMD, we still utilize the model assumed in (13) to denoise the image patches, which can be approximated by the combination of few nonzero atoms in the dictionary $\mathbf{B}\tilde{\mathbf{D}}_0$. The denoised image $\hat{\mathbf{Y}}$ is obtained by averaging the denoised patches at their respective locations.

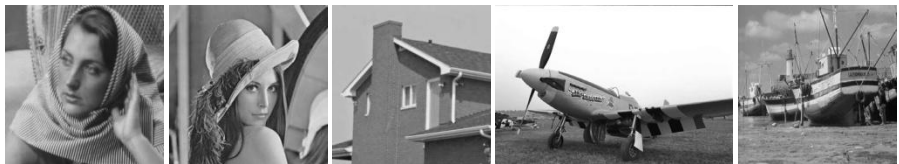

**Fig. 6.** Training images used for learning dictionaries.

To make a fair comparison, we keep the same dimension setting as the K-SVD and OCTOBOS methods, where the patch size is fixed at $8 \times 8$ and the size of the learned dictionary is $256 \times 64$. The means of the patches are removed during optimization and we only sparsify the mean-subtracted patches. The means are added back to the final denoised patches. The training patches are extracted from the image to be denoised or from 5 clean typical natural images, which are presented in Fig. 6. The dictionary is learned from 20000 patches randomly chosen from the input image. For K-SVD and OCTOBOS methods, the dictionary is initialized using an overcomplete DCT dictionary. The proposed algorithm can guarantee finding a critical point of the relating nonconvex problem without the need of initial dictionary.

In this experiment, we have qualitatively tested the denoising effectiveness on image *Cameraman* and *Airplane*, from which the training image patches including noise are chosen. See Figs. 7 and 8 for the visual illustration of the denoising results. It is evident that the proposed method outperforms other three schemes with the better perceptual quality and less artifacts. Moreover, our scheme can better preserve the textures and patterns of images and recover more clear details at intermediate and high noise levels. This could be explained by the fact that the proposed multiscale dictionary structure and added sparsity in a cooperative manner can effectively capture edges or textures of natural images and the IMF atoms get more adapted to image-specific features. The results suggest that highly-decorrelated sparsity constraint can decrease the scale mixing with the strong noise and the REMD can prevent overfitting to noise.

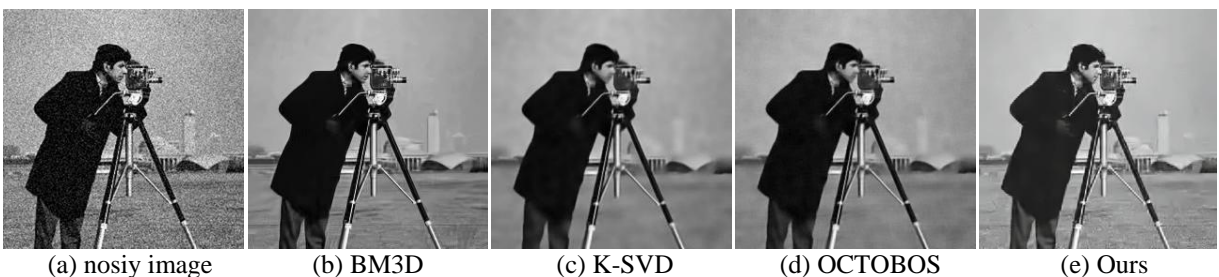

(a) nosiy image    (b) BM3D    (c) K-SVD    (d) OCTOBOS    (e) Ours

**Fig. 7.** Denosing results for image *Cameraman* with Gaussian noise $\sigma = 20$.

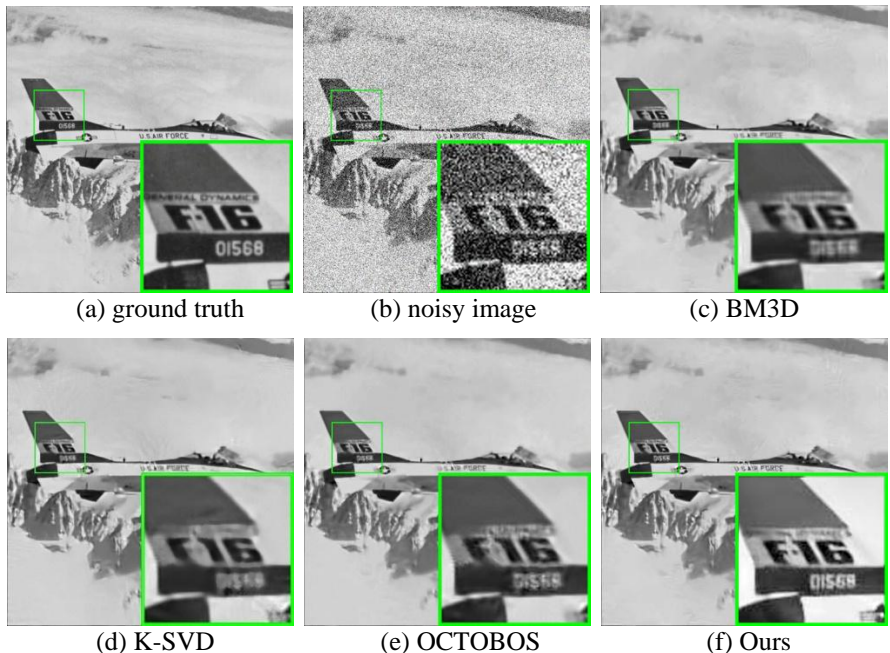

(a) ground truth    (b) noisy image    (c) BM3D

(d) K-SVD    (e) OCTOBOS    (f) Ours

**Fig. 8.** Denosing results for image *Airplane* with Gaussian noise $\sigma = 60$.

The convergence behaviors of the comparable learning methods are illustrated in Fig. 9 at different noise levels. It can be seen that our method noticeably reduces the objective function values of (14) and (17) faster than OCTOBOS method. It also shows that practical convergence properties of the proposed fast proximal scheme agree with the theoretical analysis in section 4.3. The denoising schemes are quantitatively tested on 4 natural images with a wide noise levels. The clean images are used to train. As listed in Tables 1 and 2, the proposed scheme outperforms other three methods in most cases in terms of PSNR and SSIM because structural information distinct from the noise is efficiently extracted from training patches and injected into IMF atoms.

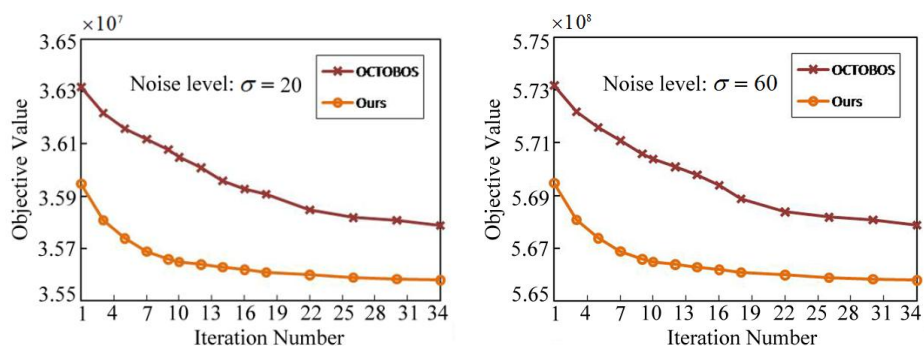

**Fig. 9.** Convergence behaviors of learning algorithms for image denoising.

**Table 1** PSNR (dB) comparison of different denoising methods.

| Image | $\sigma$ | BM3D | K-SVD | OCTOBOS | Ours |
|---|---|---|---|---|---|
| Barbara | 20 | 34.97 | 34.11 | 34.64 | **34.80** |
| | 60 | **25.61** | 24.01 | 24.41 | 25.27 |
| Lena | 20 | 35.78 | 34.92 | 35.26 | **35.89** |
| | 60 | 26.49 | 25.62 | 26.05 | **26.74** |
| Cameraman | 20 | 34.15 | 33.82 | 33.83 | **34.26** |
| | 60 | 25.82 | 23.96 | 24.84 | **25.83** |
| Airplane | 20 | 31.81 | 33.32 | 31.52 | **31.92** |
| | 60 | **25.67** | 24.57 | 25.19 | 25.30 |

**Table 2** SSIM comparison of different denoising methods.

| Image | $\sigma$ | BM3D | K-SVD | OCTOBOS | Ours |
|---|---|---|---|---|---|
| Barbara | 20 | 0.9128 | 0.9049 | 0.9150 | **0.9174** |
| | 60 | 0.7426 | 0.7560 | **0.7587** | 0.7536 |
| Lena | 20 | 0.9069 | 0.9015 | 0.9041 | **0.9188** |
| | 60 | 0.7165 | **0.7693** | 0.7611 | 0.7628 |
| Cameraman | 20 | 0.9136 | 0.9159 | 0.9178 | **0.9205** |
| | 60 | 0.7634 | 0.7571 | 0.7627 | **0.7710** |
| Airplane | 20 | 0.8534 | 0.8522 | 0.8526 | **0.8545** |
| | 60 | 0.7893 | 0.7874 | 0.7897 | **0.7923** |

## 6. Conclusion

In this paper, we propose a novel multiscale dictionary learning method for sparse image representation based on the improved EMD. By using this fully-adaptive data-driven decomposition tool, image data can be sparsely represented at a relatively low sparsifying level. For leading to more compact and effective representation, a set of IMF atoms are selected from the raw IMFs via cross-scale frequency clustering. Besides, a tolerance dictionary is adjoint to further promote sparsity and generalization. The multiplication of an IMF dictionary and a tolerance dictionary is employed to find physical meaning representation and capture the spatial, scale and directional characteristics of images. In addition, row coherence is introduced as prior information to penalize the correlations between the atoms. We use a fast proximal scheme to solve the coherence regularized model. Extensive experiments on natural image have confirmed the competitive performance of the proposed learning algorithm as compared with other representative methods. The applications in image denoising task have also shown the merits of our synthesized multiscale dictionary involving minimum sparsifying error and high visual quality.

## Acknowledgments

This research was supported by China Postdoctoral Science Foundation (2014M560852) and Major National Scientific Instrument and Equipment Development Project of China (2013YQ030967).